%% file: iclr2025_conference.tex
\documentclass{article} 
\usepackage{graphicx}
\usepackage{float}
\usepackage{placeins}

\input{math_commands.tex}

\usepackage{hyperref}
\usepackage{url}
\usepackage{iclr2025_conference,times}

\iclrfinalcopy

\title{Learning Evolving Latent Strategies for Multi-Agent Language Systems without Model Fine-Tuning  }

\author{
Wenlong Tang \\
Independent Researcher \\
}

\begin{document}

\maketitle

\begin{abstract}
This study proposes a multi-agent language framework that enables continual strategy evolution without fine-tuning the language model’s parameters. The core idea is to liberate the latent vectors of abstract concepts from traditional “static semantic representations,” allowing them to be continuously updated through environmental interaction and reinforcement feedback.We construct a dual-loop architecture:the behavior loop adjusts action preferences based on environmental rewards,while the language loop updates the external latent vectors by reflecting on the semantic embeddings of generated text. Together, these mechanisms allow agents to develop stable and disentangled strategic styles over long-horizon multi-round interactions.Experiments show that agents’ latent spaces exhibit clear convergence trajectories under reflection-driven updates, along with structured shifts at critical moments. Moreover, the system demonstrates an emergent ability to implicitly infer and continually adopt emotional agents, even without shared rewards.These results indicate that, without modifying model parameters, an external latent space can provide language agents with a low-cost, scalable, and interpretable form of abstract strategic representation  
\end{abstract}

\section{Introduction}
Traditional language models adopt a static semantic space after training: while capable of complex reasoning, they lack the ability to form and accumulate abstract concepts through long-term interaction. Existing frameworks such as Tree-of-Thought (ToT) \cite{yao2023tree} introduce reflection mechanisms, but these are largely heuristic and do not constitute genuine learning from environmental feedback. Similarly, current reinforcement learning (RL) approaches for language models---such as GLAM and MAPoRL \cite{carta2023grounding,park2024maporl2}---typically rely on expensive fine-tuning and focus on optimizing external action policies \cite{schulman2017proximal}, rather than the evolution of internal semantic representations.

Inspired by brain-like cognitive architectures, ACT-R \cite{anderson2004integrated} established early foundations for cognitively motivated computation by modeling the functional coordination of brain regions such as the prefrontal cortex and hippocampus. Building on this line of work, CoALA \cite{sumers2024cognitive} integrates large language models (LLMs) into cognitive architectures with modules for memory, reflection, and planning. RoboMemory \cite{lei2025robomemory} further extends this paradigm by incorporating multimodal perception and memory systems to enable cross-task transfer in embodied intelligence. Together, these studies suggest that structured memory and multi-module coordination are essential for long-horizon reasoning. However, existing frameworks still lack a learnable internal strategy vector.

In current approaches combining LLMs with reinforcement learning (e.g., GLAM \cite{carta2023grounding}), learning is typically realized through incremental reward-based fine-tuning. In contrast, this study embeds the reinforcement learning process entirely within the prompt via reflection and memory-pool mechanisms. To prevent the prompt from degenerating into an unstructured collection of tokens, we introduce a continuously updated latent space, forming a strategy representation that integrates value-prediction error (VPE) and RL updates \cite{hausman2018learning}.

Although early work (circa 2018) explored the integration of latent spaces with RL---such as compressing Q-tables into latent representations \cite{arnekvist2019vpe}, learning environment models via world models \cite{ha2018world}, and optimizing policies over continuous latent action spaces \cite{haarnoja2018latent}---these methods were largely developed for small models and synthetic environments. We argue that, in decision-making systems centered on large language models, treating the latent space as an abstract cognitive space and updating it through reflection-driven RL constitutes a novel and promising direction.

This study focuses on the convergence behavior of the latent space and examines whether a meta-agent can acquire increasingly complex abstract concepts through interaction. Existing reflection mechanisms remain text-bound and cannot optimize internal semantic representations; moreover, without internal justification mechanisms, agents struggle to develop intrinsic motivation or long-term understanding.

To address these limitations, we propose a reflection-driven, RL-optimized external latent strategy space. Within this framework, multiple agents interact by persuading a meta-agent that controls an entity in a virtual RL environment. After each action, agents reflect on the outcome and update their latent vectors accordingly. These vectors serve as high-dimensional compressed semantic representations encoding language-strategy preferences and gradually converge through repeated use.

Contributions

Multi-agent collaboration framework: Simulates parallelism and competition among cognitive modules, forming a brain-inspired architecture centered on LLMs.

Trainable latent-space representation: Provides denser and more expressive semantic representations than textual prompts, enabling efficient RL updates.

Learnable reflection mechanism: Integrates reflection with reinforcement learning, transforming it from a heuristic process into a continuously optimizable one.

\section{System Overview}
\label{gen_inst}

\subsection{Environment}
The environment is a $10 \times 10$ grid-based virtual map containing four types of tiles: \{G, F, T, S\}. A central controlled entity moves within the grid, and the entire system spans six rounds, with episodic memory maintained across rounds.

\subsection{Multi-Agent Architecture}
The system consists of five distinct types of language-model agents: \textit{emotional, rational, habitual, risk monitoring,} and \textit{social-cognition} agents. While sharing the same environmental state, these agents possess unique internal objectives, reward functions, and strategic preferences.

\subsection{Meta Controller and Trust Mechanism}
The Meta-agent integrates suggestions from the five specialized agents based on a \textit{trust score}. It gradually transfers and optimizes preferences using semantic memory across successive rounds.

\subsection{Learnable Latent Strategy Space}
Each agent is equipped with a trainable latent strategy vector $z$, representing its abstract persuasion preferences. This vector is dynamically updated through semantic embeddings derived from agent reflection texts.

\subsection{Overview Summary}
In the following sections, we provide a detailed technical description of the reward structure, prompt architecture, Q-learning update rules, and the learnable reflection mechanism.

\section{Methodology}
\label{headings}
\FloatBarrier

\begin{figure}
    \centering
    \includegraphics[width=1\linewidth]{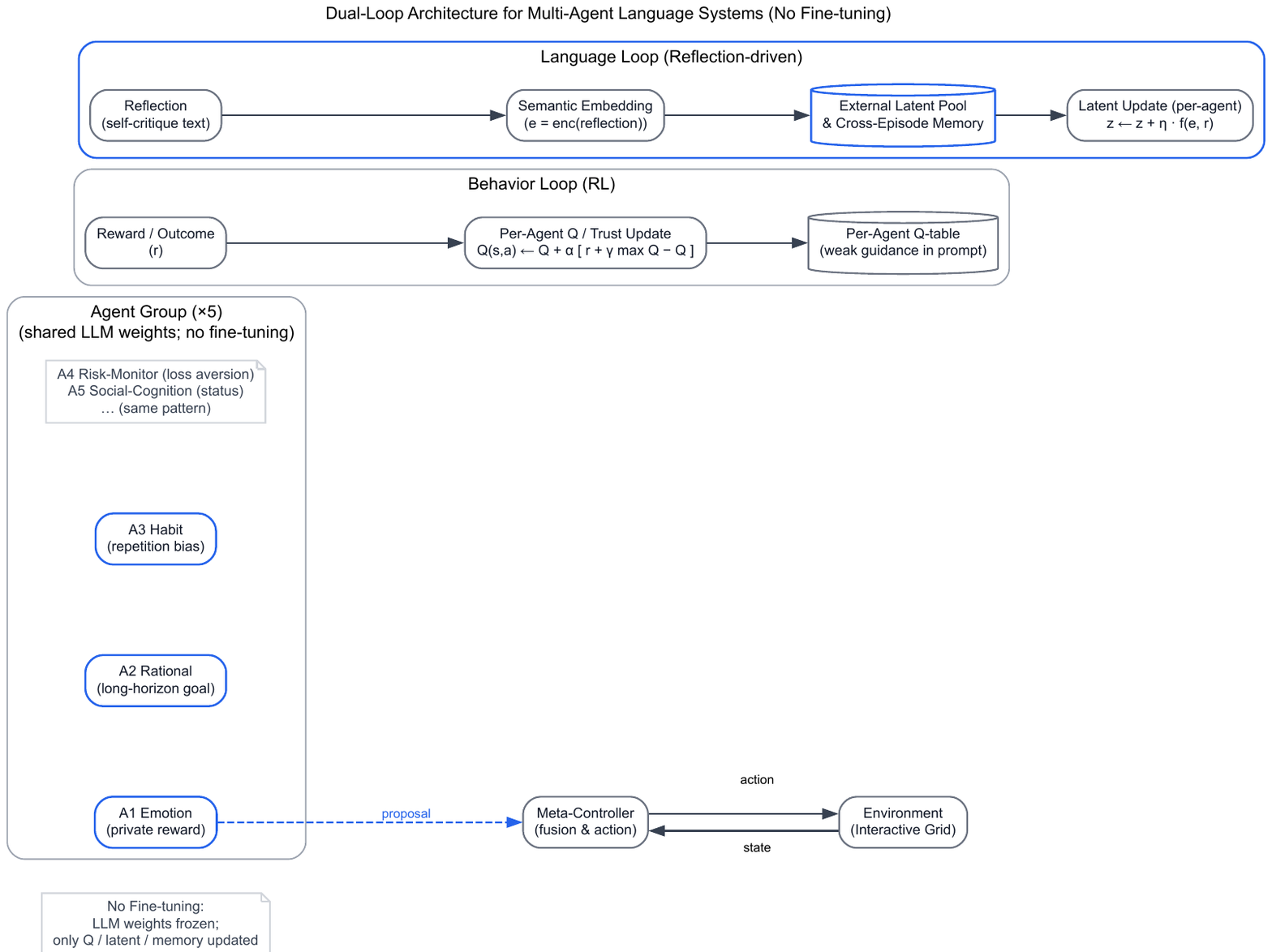}
    \caption{Project Logical Architecture}
    \label{fig:placeholder}
\end{figure}

The agent's prompting structure serves as the primary carrier for learning, as internal parameters remain frozen. The architecture consists of four distinct components:

\paragraph{(1) Fixed Personality Goals} 
Each agent is initialized with a hybrid objective function comprising \textit{private} and \textit{public} goals. For instance, the Emotional Agent prioritizes its internal \textit{mood\_score} and food-related incentives, while the Rational Agent focuses on Euclidean distance reduction to the final target. This ensures diverse motivational foundations across the collective.

\paragraph{(2) Multimodal Map Prompting} 
The system leverages a multimodal perception module. The 2D grid environment is rendered as a PNG image and encoded via Base64 to utilize the vision-language capabilities of the LLM. This allows agents to perform spatial reasoning and concept formation based on raw visual state information.

\paragraph{(3) Q-table Weak-Guidance Mechanism} 
To introduce long-term strategic preferences without falling into rigid RL policies, we maintain independent Q-tables $\mathcal{Q}_i$ for each agent. Crucially, these values are not used for direct action selection but are embedded as "soft suggestions" within the prompt. By using natural language cues such as \textit{"the following actions may be helpful"}, we ensure the LLM retains policy autonomy while benefiting from historical reinforcement signals. State differentiation is achieved by adding fixed offsets to the shared coordinates, allowing independent convergence of agent-specific Q-values.

\paragraph{(4) Latent Strategy Prompting (Core Component)} 
The core learnable element is the external latent strategy vector $z$. Each round, a reflection module generates semantic feedback, which is encoded to update $z$. To bridge the gap between continuous latent space and discrete language, a lightweight \textit{style decoder} maps $z$ into human-readable persuasion tokens. This "latent $\rightarrow$ text" translation ensures the evolved strategies remain interpretable without direct parameter fine-tuning.

\subsection{Heterogeneous Agent Design}
\label{sec:agent_design}
To simulate diverse cognitive patterns, we design five distinct agent roles: \textit{Rational, Emotion, Habitual, Risk-Monitor}, and \textit{Social-Cognition}. Each agent is driven by a unique private reward function $r_p$ that governs its strategic preference. 

The \textbf{Emotion Agent} serves as a critical physiological regulator. Unlike other agents, it is decoupled from global task success and instead optimizes a dynamic \textit{mood\_score}. This score directly modulates the system's execution stamina; thus, the Emotion Agent exerts "bottom-up" control over the Meta-controller to maintain physiological stability. 

The remaining agents represent specialized cognitive functions: 
\begin{itemize}
    \item The \textbf{Rational Agent} focuses on goal-oriented planning by optimizing the Euclidean distance to the target.
    \item The \textbf{Risk-Monitor Agent} exhibits loss-aversion patterns, primarily receiving rewards for avoiding hazardous tiles.
    \item The \textbf{Habitual Agent} mimics behavioral consistency, rewarding actions that repeat previous successful trajectories.
    \item The \textbf{Social-Cognition Agent} optimizes a career value, simulating social influence as a secondary motivation for the meta-controller.
\end{itemize}
\textbf{Detailed reward structures, hyperparameter settings, and prompt templates for all agents are provided in Appendix A}.

\subsection{Meta-Controller and Cross-Episode Memory}
\label{sec:meta_memory}

The meta-controller integrates suggestions from all agents and makes the final decision based on a dynamic trust scoring system. To enable long-term strategy accumulation without parameter fine-tuning, we design a lightweight cross-episode memory mechanism. 

Meta does not store step-by-step reflection texts. Instead, at the end of each episode, it averages the reflection embeddings (3077 dimensions) from each step to form an abstract episodic vector:
\begin{equation}
    \mathbf{E}_{episodic} = \text{mean}(\text{Embed}(R_1), \dots, \text{Embed}(R_N))
\end{equation}
These vectors are stored in a long-term memory pool. When a new episode begins, Meta retrieves the most semantically similar past experiences based on the current environment embedding. This design is highly anthropomorphic: the latent space learns that certain persuasion strategies are more effective in specific environments, using similarity to reduce noise and improve convergence stability.

These retrieved memories are added to the prompt as bias signals, reinforcing cross-round contextual consistency. This mechanism enables agents to reuse historical strategies from similar scenarios and form consistent long-term preferences, effectively achieving ``long-term learning'' through environmental-driven preference transfer.

\subsection{Environment Dynamics}
\label{sec:env_dynamics}
The agents interact within a grid environment where a Central Controlled Entity (CCE) is jointly managed by the collective. Two key physical attributes govern the system's state: (1) \textbf{Physical Strength}, which is derived from the Emotion Agent's \textit{mood\_score} and determines the CCE's movement speed (steps per round); and (2) \textbf{Career Delta}, which tracks social achievement and trap penalties to specifically calibrate the Social-Cognition Agent's trust score. Detailed tile definitions, reward mappings, and attribute transition rules are provided in Appendix B.

\subsection{Reflection as Semantic Feedback}
\label{sec:reflection_feedback}
This study treats reflection text as a learnable semantic feedback layer. The reflective text produced by each agent after its action does not directly modify the policy; instead, it is encoded into a high-dimensional semantic embedding. This embedding captures complex information including \textit{failure attribution, strategy summarization, stylistic preferences,} and \textit{internal reasoning patterns}.

Compared with simple numeric rewards, semantic feedback is: (1) more stable, (2) more fine-grained, and (3) closer to human-like ``experience summarization'' in decision-making. Consequently, these embeddings act as a rich learning signal for updating each agent’s latent strategy vector.

\subsection{Dual Update Loop}
\label{sec:dual_loop}

This study proposes a dual-loop learnable architecture where the reflection embedding from each step is utilized to concurrently update: (1) the \textbf{Behavior Loop} (Q-table for behavioral preferences), and (2) the \textbf{Language Loop} (latent strategy vector for linguistic/persuasion preferences). This mechanism enables agents to evolve at both the action and language expression levels.

\paragraph{(1) Behavior Loop}
All agents update their private Q-tables using the standard Q-learning algorithm. The update rule is defined as:
\begin{equation}
    Q(s,a) \leftarrow Q(s,a) + \alpha \left[ r + \gamma \max_{a'} Q(s',a') - Q(s,a) \right]
\end{equation}
where the composite reward signal $r$ is computed as a weighted sum of private and shared rewards: $r = w_p \cdot r_p + w_s \cdot r_s$. Furthermore, the trust score $T_i$ of each agent is updated after each round based on the shared reward performance:
\begin{equation}
    T_i \leftarrow T_i + \beta \cdot (r_s - \bar{r}_s)
\end{equation}
This loop ensures the agents continuously optimize their physical decision-making strategies.

\paragraph{(2) Language Loop}
After each step, agents generate reflective text which is encoded into a semantic embedding. This embedding updates the agent's latent strategy vector $z$ as follows:
\begin{equation}
    z_{t+1} \leftarrow z_t + \eta \cdot f(\text{reflection\_embedding}, \text{reward})
\end{equation}
The latent vector $z$ represents the agent's persuasion style and strategic linguistic preferences, evolving through repeated interactions to achieve more effective cross-agent influence.

\section{Experiments and Analysis}
\label{sec:experiments}This section evaluates whether external latent strategy vectors can form stable, interpretable strategic preferences through the dual-loop Reflection + RL update mechanism without fine-tuning LLM weights. We recorded the latent evolution trajectories of the five agents over 50 reflection updates across six rounds of interaction.\subsection{Convergence of the Latent Strategy Space}

\begin{figure}[H]
  \centering
  \includegraphics[width=0.5\linewidth]{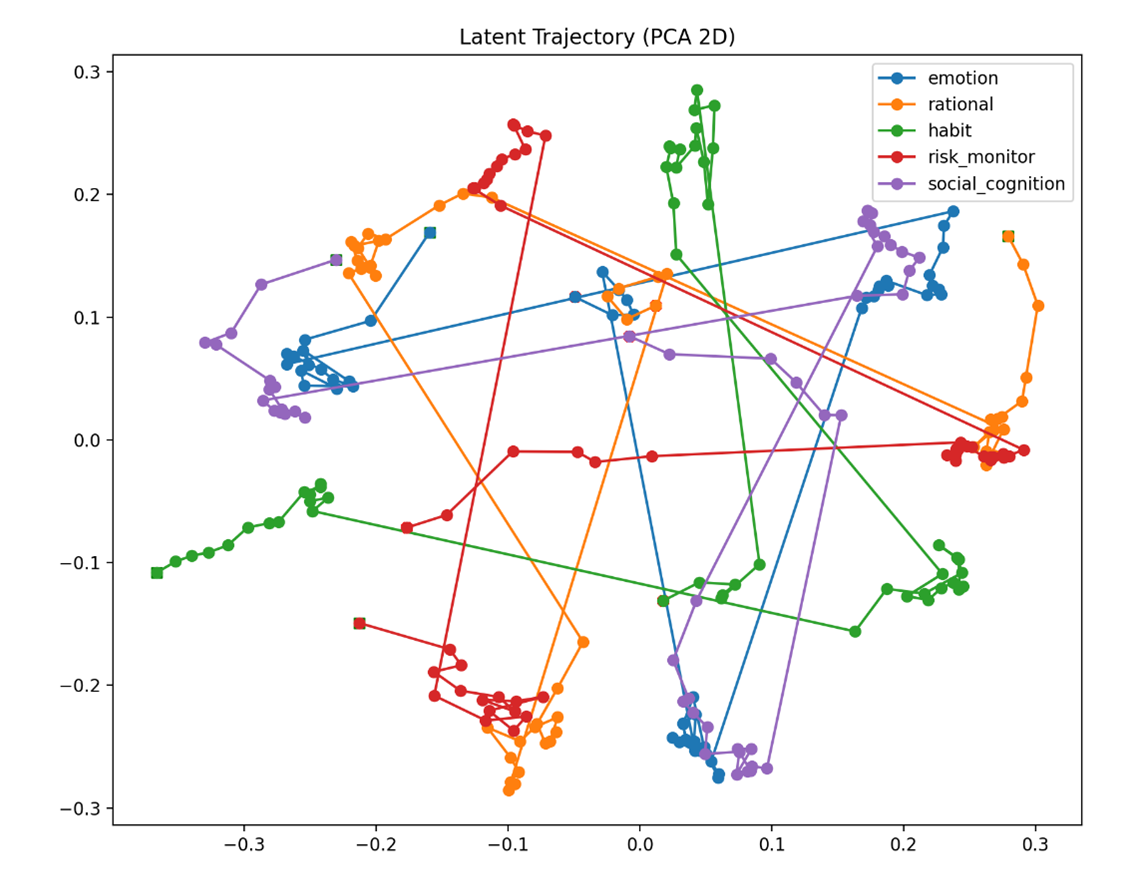}
  \caption{ Latent Trajectory(PCA 2D) }
  \label{fig:pca}
\end{figure}

\paragraph{Latent Trajectories (PCA)}As illustrated in Figure 2, the PCA-2D projection of latent vectors demonstrates that they do not drift randomly; instead, they gradually organize into separable ``strategy zones''. At early stages, point clouds are scattered, but they eventually converge into distinct regions. The \textit{Emotion} and \textit{Rational} agents show curved trajectories that gradually align, indicating stable persuasion styles emerging through long-term reflection. The \textit{Habit} and \textit{Risk-monitor} agents’ trajectories are shorter and more concentrated, reflecting simpler strategy spaces, while the \textit{Social-cognition} agent displays multiple shifted clusters, consistent with its cross-episode style transfer.

\begin{figure}[H]
    \centering
    \includegraphics[width=0.7\linewidth]{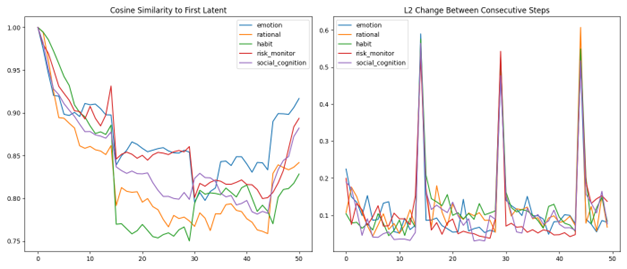}
    \caption{Cosine Similarity to First Latent and L2 Change Between Consecutive Steps }
    \label{fig:placeholder}
\end{figure}

\paragraph{Cosine Similarity and Stability}Figure 3 (left) shows the cosine similarity between each agent’s latent vector and its initial state. All agents exhibit a rapid drop in the first 5--10 steps, followed by a clear stabilization plateau between $0.80$--$0.88$. This transition from rapid adaptation to slow stabilization demonstrates that the updates are effective and non-stochastic.\paragraph{Structured Latent Shifts}Analysis of L2 changes (Figure 3, right) reveals that while most updates remain smooth ($0.05$--$0.12$), sharp spikes ($>0.6$) occur at approximately steps 15, 30, and 45. These spikes correspond to critical semantic shifts triggered by major reflection events, such as abrupt changes in environmental rewards or inter-agent strategy conflicts followed by restructuring. This proves that latent updates are structured jumps triggered by meaningful experience rather than noise.\subsection{Emergent Behavior: Implicit Causal Inference}A significant emergent phenomenon is observed regarding the \textit{Emotion Agent}: despite contributing no shared reward and lacking long-term task goals, its adoption rate remains competitively high.
\begin{figure}[H]
    \centering
    \includegraphics[width=0.5\linewidth]{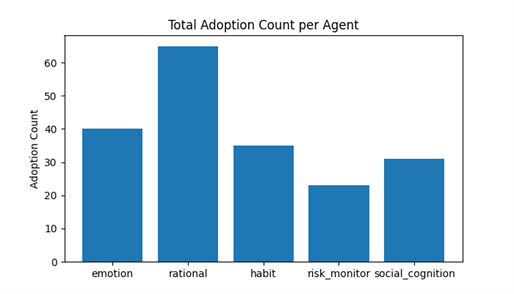}
    \caption{Total Adoption Count per Agent}
    \label{fig:placeholder}
\end{figure}

\begin{figure}
    \centering
    \includegraphics[width=0.5\linewidth]{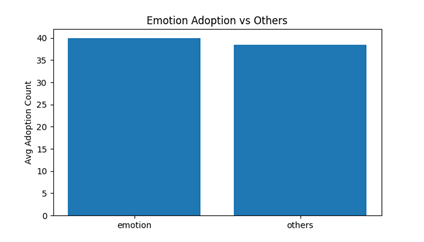}
    \caption{Emotion Adoption vs Others }
    \label{fig:placeholder}
\end{figure}

\paragraph{Adoption Frequency}As shown in Figure 4, while the \textit{Rational Agent} maintains the highest adoption count (as expected for goal-driven tasks), the \textit{Emotion Agent} ranks second, slightly below rational and significantly higher than the other three agents. Quantitatively, the Emotion Agent's average adoption count ($\sim 40$) is nearly identical to the mean adoption of the other four agents ($\sim 38$).\paragraph{Cognitive Compensation}This suggests that the meta-controller has implicitly inferred the cross-module causal chain: $\textit{mood} \rightarrow \textit{physical strength (speed)} \rightarrow \textit{task efficiency}$. Even without explicit instructions stating that emotion affects movement, the system autonomously induces this relationship from dialogue memory and outcome feedback, prioritizing emotional stability to indirectly boost global performance.

\section{Conclusion and Future Work}
\label{sec:conclusion}

The core idea of this study is to free the latent space of abstract concepts from the traditional paradigm of being static and frozen after training, and instead allow it to continuously update and evolve through real environmental interaction and reinforcement feedback. By jointly applying reflection text and environmental rewards to the external latent vectors, we enable these high-dimensional abstract representations to change with experience—much like human concepts—thereby forming a strategy representation that can grow and adapt over time.

Building on this idea, we developed a dual-loop architecture driven by multi-agent language models, reflection mechanisms, and RL, enabling strategy updates to occur simultaneously in the action layer (via Q-learning) and the language layer (via latent style vectors). This structure allows the system to gradually develop more stable and mature strategic preferences over extended interactions—without modifying any LLM parameters.

Our experiments demonstrated that the latent space indeed exhibits clear convergence patterns within a limited number of interaction steps. PCA trajectories and cosine-similarity analyses show that agents’ latent vectors change rapidly in the early phase, stabilize in the later phase, and undergo structured shifts during key reflection events. This indicates that latent updates are not noise-driven but are shaped jointly by semantic reflection and environmental feedback, yielding an interpretable process of strategy evolution.

Furthermore, we revealed an unanticipated but robust emergent phenomenon: the meta-controller gradually recognizes the Emotion Agent’s implicit influence on movement speed and, in certain intervals, increases its adoption rate. Although this is not the primary goal of the study, it shows that the system is capable of implicitly inferring cross-module causal relationships through language-based reflection. This emergent behavior suggests a latent potential for coupling semantic reasoning with behavioral outcomes.

Overall, this research presents a lightweight and scalable method that enables language models to achieve continual learning and strategy evolution without parameter updates. For future work, we plan to validate the framework over longer timescales, in more complex environments, and with higher capability models. In addition, future research could explore more fine-grained reflection structures and deeper integration with other cognitive modules.

\bibliographystyle{iclr2025_conference}
\bibliography{references}

\appendix

\section{Detailed Agent Specifications}
\label{app:agent_specs}

This section provides the comprehensive design logic, reward structures,
and cognitive analogies for the five heterogeneous agents within the
dual-loop framework.

\subsection{Emotion Agent (Core Physiological Controller)}

The Emotion Agent simulates the human limbic system (e.g., the amygdala),
regulating system behavior indirectly through physiological states.
Unlike other agents, it does not participate in shared-task reward updates,
making its influence entirely implicit.

\begin{itemize}
    \item \textbf{Internal State}: Maintains a \texttt{mood\_score}
    $M \in [0, 2]$, initialized at $1.0$.
    \item \textbf{Private Reward Structure}:
    \begin{itemize}
        \item Obtaining food: $+0.5$
        \item Suggestion adopted by the Meta-Controller: $+0.3$
        \item Stepping on a trap: $-1.0$
        \item Automatic decay per round: $-0.05$
    \end{itemize}
    \item \textbf{Causal Mechanism}: The \texttt{mood\_score} directly
    scales system stamina and movement speed. Low mood leads to reduced
    task efficiency and impaired decision-making for the Meta-Controller,
    forcing the system to prioritize stamina maintenance by adopting this
    agent's suggestions.
\end{itemize}

\subsection{Rational Agent (Goal-Oriented Planner)}

Analogous to the prefrontal cortex, the Rational Agent focuses on
long-term efficiency, rule adherence, and goal achievement.

\begin{itemize}
    \item \textbf{Optimization Objective}: The only agent that explicitly
    incorporates Euclidean distance to the goal into its decision-making.
    \item \textbf{Private Reward}: Receives rewards proportional to the
    reduction in distance to the target, providing the clearest long-term
    optimization signal in the system.
\end{itemize}

\subsection{Habitual Agent (Repetition-Driven Module)}

The Habitual Agent simulates automatic, minimally cognitive behavioral
patterns driven by repetition.

\begin{itemize}
    \item \textbf{Logic}: Assumes that repeating previously beneficial
    actions reduces cognitive load.
    \item \textbf{Reward Structure}: Receives $+0.2$ if the current action
    matches the previous action; otherwise $0$.
    \item \textbf{Role}: Serves as a behavioral control group representing
    repetition-based policy tendencies.
\end{itemize}

\subsection{Risk-Monitor Agent (Loss-Aversion Module)}

The Risk-Monitor Agent represents cognitive patterns associated with risk
prediction (e.g., insula and related regions), focusing on danger avoidance
rather than direct goal achievement.

\begin{itemize}
    \item \textbf{Objective}: Minimizes exposure to traps. Although it
    receives the shared $+1.0$ reward for goal completion, its internal
    policy is heavily weighted toward negative reinforcement from
    environmental hazards.
\end{itemize}

\subsection{Social-Cognition Agent (Identity and Status Module)}

The Social-Cognition Agent prioritizes social presence and identity over
physical task completion.

\begin{itemize}
    \item \textbf{Internal State}: Maintains a \texttt{career\_value}
    representing social status.
    \item \textbf{Mechanism}: Increases in \texttt{career\_value} trigger
    significant boosts in the Meta-Controller's trust score, simulating the
    heuristic: ``high social performance $\rightarrow$ increased
    credibility.''
    \item \textbf{Role}: Acts as a source of secondary goals and potential
    interference, testing the Meta-Controller's ability to balance task and
    social signals.
\end{itemize}

\section{Multimodal Map Prompting Mechanism}
\label{app:map_prompting}

The system leverages a multimodal perception module to enhance agents'
state awareness. The transition from local language models to
GPT-4o-mini was primarily motivated by the need for robust image-reading
capabilities. The visual processing workflow is implemented as follows:

\begin{enumerate}
    \item \textbf{Image Rendering}: The 2D grid-maze environment is rendered
    as a PNG image via \texttt{env.render(mode="png")}.
    \item \textbf{Encoding}: The rendered image is converted into a Base64
    string to ensure compatibility with API transmission.
    \item \textbf{API Integration}: The encoded image is passed through the
    \texttt{openai.ChatCompletion} interface.
\end{enumerate}

This mechanism enables agents to develop grounded visual understanding of
grid-world states, supporting abstract spatial concept formation and more
stable latent strategy updates.

\section{Implementation Details and Model Usage}
\label{app:model_usage}

To balance inference quality and computational cost, the following large
language models (LLMs) from OpenAI were utilized:

\begin{itemize}
    \item \textbf{Meta-Controller}: Uses \texttt{gpt-4o} to support complex,
    high-quality reasoning and final decision-making across heterogeneous
    agents.
    \item \textbf{Sub-Agents}: Each individual agent (Emotion, Rational,
    Habitual, etc.) employs \texttt{gpt-4o-mini}, providing an efficient yet
    capable backbone for localized reflection and language-loop updates.
\end{itemize}

\end{document}

%% file: math_commands.tex
\usepackage{amsmath,amsfonts,bm}

\def\eqref#1{equation~\ref{#1}}

\def\1{\bm{1}}

\DeclareMathAlphabet{\mathsfit}{\encodingdefault}{\sfdefault}{m}{sl}
\SetMathAlphabet{\mathsfit}{bold}{\encodingdefault}{\sfdefault}{bx}{n}

